\documentclass[10pt,twocolumn,letterpaper]{article}

\usepackage[pagenumbers]{cvpr} 
\usepackage{graphicx}
\usepackage{multirow}
\usepackage{amsmath}
\usepackage{booktabs}
\usepackage{graphicx}
\usepackage{natbib}
\setcitestyle{square, comma, numbers,sort&compress, super}
\usepackage{float}
\usepackage{amsthm}
\usepackage{graphicx}
\usepackage{epstopdf}
\usepackage{color}
\usepackage{soul}
\usepackage{verbatim}
\usepackage{tikz,times}
\usepackage{multirow}
\usepackage{gensymb}
\usepackage{lipsum} 
\usepackage{color, colortbl}

\definecolor{cvprblue}{rgb}{0.21,0.49,0.74}
\usepackage[pagebackref,breaklinks,colorlinks,allcolors=cvprblue]{hyperref}

\title{
AgentAlign: Misalignment-Adapted Multi-Agent Perception for Resilient Inter-Agent Sensor Correlations
}

\author{{Zonglin Meng, Yun Zhang, Zhaoliang Zheng, Zhihao Zhao, Jiaqi Ma} 
\thanks{Zonglin Meng, Yun Zhang, Zhaoliang Zheng, Zhihao Zhao, Jiaqi Ma are affiliated with the University of California, Los Angeles,
CA 90095 USA.}
}

\begin{document}
\maketitle
\begin{abstract}
Cooperative perception has attracted wide attention given its capability to leverage shared information across connected automated vehicles (CAVs) and smart infrastructures to address sensing occlusion and range limitation issues. However, existing research overlooks the fragile multi-sensor correlations in multi-agent settings, as the heterogeneous agent sensor measurements are highly susceptible to environmental factors, leading to weakened inter-agent sensor interactions. The varying operational conditions and other real-world factors inevitably introduce multifactorial noise and consequentially lead to multi-sensor misalignment, making the deployment of multi-agent multi-modality perception particularly challenging in the real world. In this paper, we propose AgentAlign, a real-world heterogeneous agent cross-modality feature alignment framework, to effectively address these multi-modality misalignment issues. Our method introduces a cross-modality feature alignment space (CFAS) and heterogeneous agent feature alignment (HAFA) mechanism to harmonize multi-modality features across various agents dynamically. Additionally, we present a novel V2XSet-noise dataset that simulates realistic sensor imperfections under diverse environmental conditions, facilitating a systematic evaluation of our approach's robustness. Extensive experiments on the V2X-Real and V2XSet-Noise benchmarks demonstrate that our framework achieves state-of-the-art performance, underscoring its potential for real-world applications in cooperative autonomous driving. The controllable V2XSet-Noise dataset and generation pipeline will be released in the future.
\end{abstract}    
\section{Introduction}
\label{sec:intro}
Precisely perceiving the complex driving environment is crucial for the safety of autonomous vehicles (AVs) and intelligent transportation systems (ITS) \cite{10443037, OS}. With advancements in deep learning and computer vision, environmental perception—the cornerstone of autonomous driving—has made significant strides in object detection and segmentation tasks \cite{xia2023automated, 10225497, rs11212492, 10507865, 3365235}. These advancements have led to substantial improvements in downstream tasks, including object tracking, trajectory prediction, and motion planning and control. However, the complexity of traffic scenarios and varying physical conditions with severe occlusions present challenges in achieving robust and safe sensing performance using only an individual vehicle's view. This limitation has sparked a growing interest in multi-agent information fusion to enhance perception capabilities in both academic and industrial settings.

Recently, vehicle-to-everything (V2X) \cite{10148929, xu2021opencda, xu2023opencda, xiang2024v2x} cooperative perception has emerged as a powerful approach to share multi-agent sensor measurements for a complete and accurate understanding of the environment. Existing cooperative perception methods usually follow an intermediate feature fusion pipeline through inter-agent feature combinations \cite{NEURIPS2022_1f5c5cd0, wang2020v2vnet, Li_2021_NeurIPS, Li_2021_RAL, liu2020when2com, 10160871}. However, such methods often leverage a single type of sensor and easily suffer from feature ambiguity and semantic deficiencies in the 3D-based methods, as well as the lack of 3D geometry information in the 2D-based approaches. To overcome the limitation of 2D and 3D sensors, existing automated driving systems (ADS) usually combine LiDARs and Cameras due to their promising and complementary sensing capabilities \cite{liu2023bevfusion}. This paper adopts these as representative sensor modalities to perceive the environment.

\begin{figure}[t]
\centering
\includegraphics[width=0.48\textwidth]{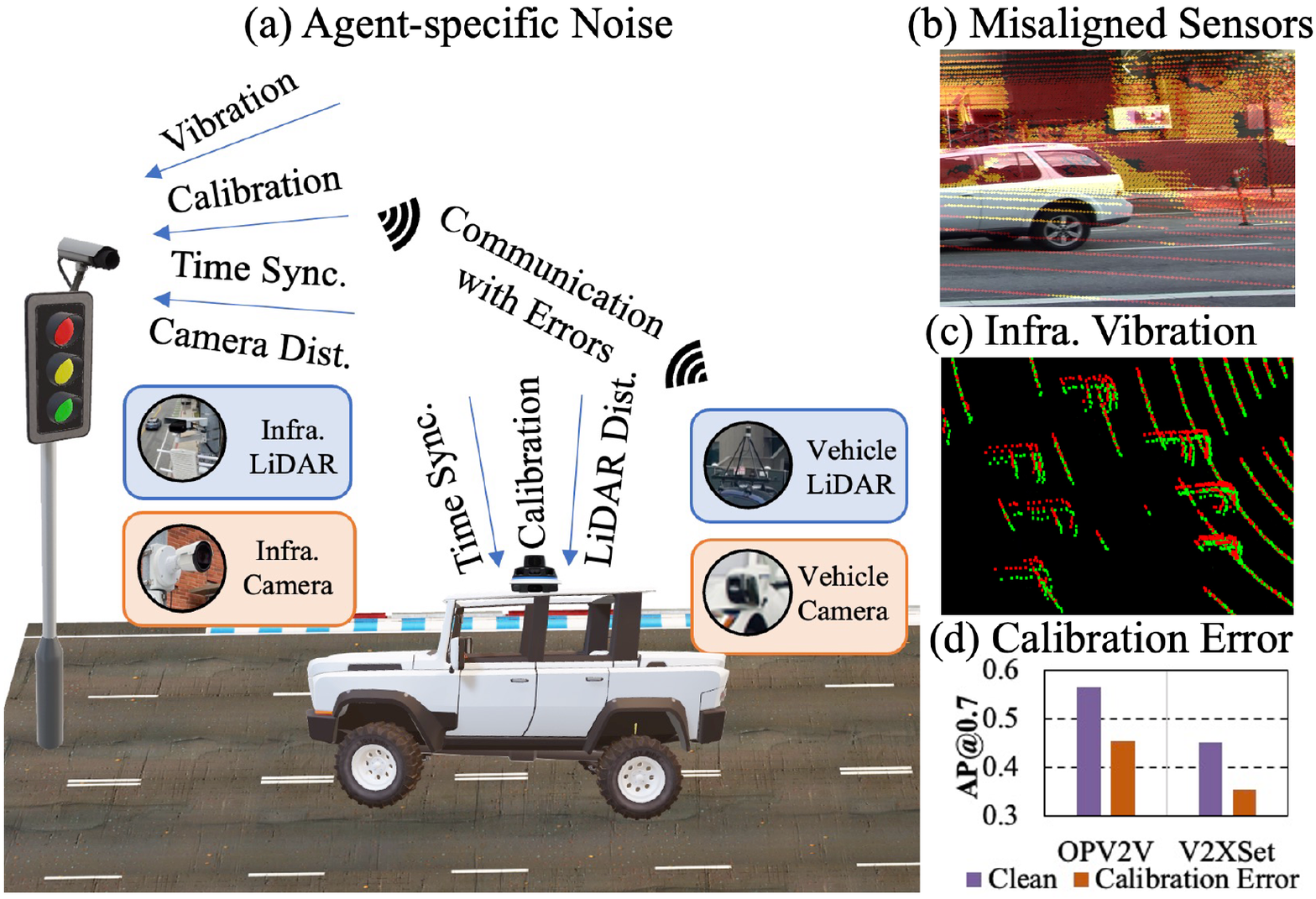}
\caption{{\bf V2X agents are typically affected by various types of noise.} In V2X settings, each agent encounters unique noise depending on its operational context as shown in Fig.(a).  In Fig.(b), we illustrate the misalignment between the camera and projected LiDAR point clouds caused by calibration errors.  Fig.(c) shows the misalignment of LiDAR point clouds on infrastructure due to vibration. In Fig.(d), a common calibration error can result in 10\% drop on AP@0.7 in both OPV2V and V2XSet, in which clean means that there are no calibration errors.}
\label{fig:problem}
\end{figure}

Although leveraging both LiDAR point clouds and camera images provides abundant features for the 3D perception tasks, integrating multiple sensor measurements in real-world multi-agent scenarios presents considerable challenges due to agent-specific misalignment issues.  
Each agent, as illustrated in \cref{fig:problem}(a), encounters distinct types of noise influenced by its operational context. Particularly, infrastructure-mounted sensors are susceptible to vibration, calibration errors, time synchronization discrepancies, and camera image distortions, which arise from exposure to environmental forces and working conditions. These factors lead to unpredictable variations in positioning and orientation, causing alignment errors in multi-modal data.
As demonstrated in \cref{fig:problem}(d), these misalignment issues lead to significant performance degradation when a random roll, yaw, or pitch angle between -0.5 and 0.5 degrees \cite{calib_2000} is introduced to the agents in V2V and V2X settings. 
We further elucidate the misalignment issues through visualizations in \cref{fig:problem}(b) and \cref{fig:problem}(c), illustrating the common errors arising from vehicle calibration and infrastructure vibration, which are usually overlooked in the current V2X perception studies. Current cooperative perception methods, usually developed with the simulated V2X dataset, directly fuse the perfect extracted features to perform the multi-agent fusion. But in the real world, the agent-specific noise and sensor multifactorial errors disrupt the coherence of multi-sensor data, impeding the network's ability to learn effective inter-agent feature interactions between different modalities. 


Therefore, to overcome the multifactorial noise caused by a variety of reasons, multi-agent alignment strategies must address two critical considerations: 1) Different agents, such as infrastructure-based and vehicle-based sensors, operate under unique conditions and are subject to multifactorial noise. Accordingly, the alignment method should adaptively harmonize the multi-modal features in a dynamic manner, accounting for the diverse errors present in agent-specific scenarios. 2) Considering the inherent sparsity of LiDAR point clouds and the distinct representation spaces of sensors, it is essential to establish a cross-modality alignment space that facilitates the seamless integration of 2D image and 3D point cloud features while overcoming the challenges posed by point cloud sparsity.

\begin{figure*}
\centering
\includegraphics[width=0.85\textwidth]{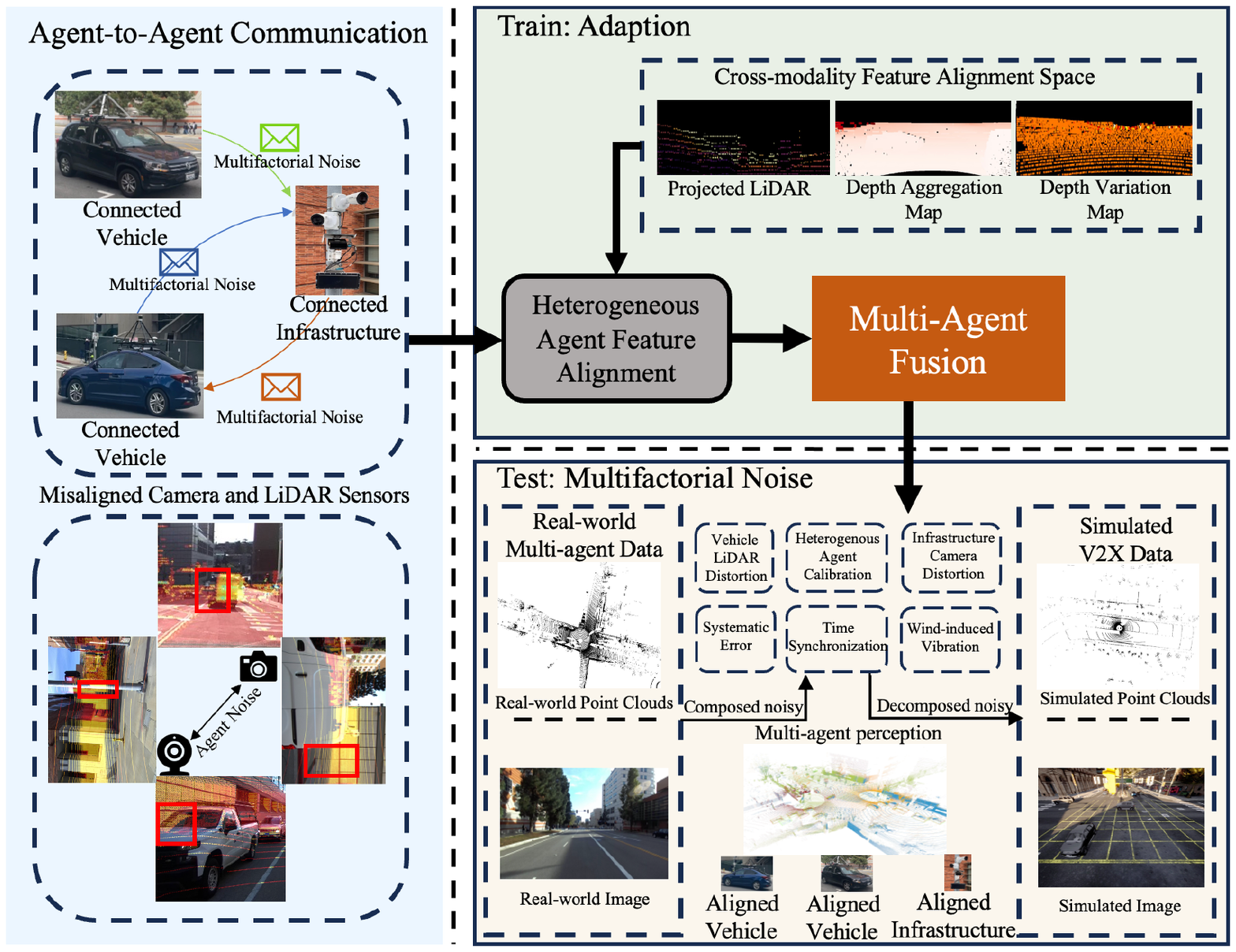}
  \caption{{\bf The detailed structure of our proposed cooperative perception framework.} In AgentAlign, connected vehicles and infrastructures communicate with each other to share the metadata and sensor measurements. Multifactorial noise often originates from agent-specific factors and propagates to other agents during communication. We propose the heterogeneous agent feature alignment method (HAFA) to dynamically align the misaligned sensor representations in the cross-modality feature alignment space (CFAS), where we construct the depth variation map from the LiDAR data for improved feature alignment. Our HAFA dynamically balances the contributions from different modality features for each unique agent. Then, each agent performs the multi-agent fusion to generate the 3D bounding boxes. We evaluate our model with real-world multi-agent data containing various types of real-world noise. Also, to systematically evaluate the impact of each type of noise, we evaluate our proposed framework on a new large-scale open dataset, namely V2XSet-Noise, to explicitly consider real-world sensor noise in V2X communication scenarios.}
\label{fig:framework}
\end{figure*}

To this end, we propose \textbf{AgentAlign}, the first-of-its-kind framework to effectively resolve real-world multi-sensor misalignment issues across heterogeneous agents. As illustrated in \cref{fig:framework}, the connected vehicles and infrastructures incorporate the proposed cross-modality feature alignment space (CFAS) and heterogeneous agent feature alignment (HAFA) method to correct, encode, and share the features with each other. The CFAS serves as the alignment space to seamlessly integrate 3D LiDAR and 2D camera features, overcoming the sparsity of the LiDAR point clouds to facilitate the learning process. Furthermore, given the multifactorial noise introduced by various agents in diverse scenarios, our HAFA dynamically accounts for agent-specific sensor misalignment, adaptively selecting informative and feature-relevant information across different modalities to improve the multi-agent sensor correlations. 


To systematically evaluate the impact of various types of noise and assess the robustness of AgentAlign, we build a new large-scale open dataset, namely V2XSet-Noise, to explicitly consider real-world sensor noise in V2X communication scenarios. This involves simulating each noise type by corrupting a purely simulated V2X cooperative perception dataset, thereby enabling a thorough analysis of our framework's performance under realistic conditions. Based on an exhaustive evaluation of both real-world and simulated noisy cooperative perception datasets, AgentAlign demonstrates robust performance by effectively addressing sensor misalignment and multifactorial noise. Our extensive experiments on the V2X-Real\cite{xiang2024v2x} and V2XSet-Noise  benchmarks demonstrate that this method establishes a new state-of-the-art performance in the field of real-world cooperative perception.

The contributions of this paper can be summarized as follows:

1. We introduce AgentAlign to resolve multi-sensor misalignment issues and capture the heterogeneous nature of V2X systems, demonstrating strong robustness against multifactorial noise and ensuring resilient inter-agent sensor correlations. Moreover, AgentAlign achieves state-of-the-art performance on complex cooperative detection benchmarks.

2. To dynamically account for the misaligned measurements and multifactorial noise arising from heterogeneous agents, we propose a cross-modality feature alignment space (CFAS) and a heterogeneous agent feature alignment (HAFA) mechanism to adaptively select and harmonize the informative features of each agent. Our approach demonstrates superior performance on both simulated and real-world multi-agent datasets, with particularly promising results on real-world noisy data, underscoring its robustness and practical applicability.

3. A novel controllable simulated V2XSet-Noise dataset and multifactorial noise generation pipeline were constructed to explicitly account for imperfect real-world multi-agent perceiving conditions. We generate vehicle and roadside infrastructure noise for both camera and LiDAR sensor measurements to reflect diverse environmental influences.

\section{Related Work}

V2X perception \cite{xu2023opencda} employs communication technology to enable autonomous vehicles (AVs) to exchange sensing data, enhancing collective environmental awareness. Early research concentrated on cooperative 3D object detection using LiDAR data, often by sharing raw point clouds (early fusion) or detection outputs (late fusion). While effective, these approaches either require considerable bandwidth or omit essential contextual details.

To achieve a balance between accuracy and latency, recent studies have explored intermediate feature sharing among neighboring vehicles, yielding notable performance improvements. OPV2V \cite{xu2021opv2v}
proposed an attentive intermediate fusion pipeline designed to effectively aggregate information from multiple connected vehicles. Where2Comm \cite{hu2022where2comm} introduces a spatial confidence map that identifies perceptually critical areas, enabling agents to share only essential information.
 V2VNet \cite{wang2020v2vnet}, for instance, introduced the use of intermediate feature sharing from 3D backbones (intermediate fusion) and applied a spatially aware graph neural network for multi-agent feature aggregation. Similarly, V2V4real \cite{10160871} uses a simple, agent-wise single-head attention mechanism for feature fusion, while F-Cooper \cite{fcooper} applies a maxout operation to combine features. DiscoNet \cite{Li_2021_RAL} incorporates knowledge distillation, aligning intermediate features with those from an early-fusion teacher model. 

For multi-modal fusion, approaches like HM-ViT \cite{xiang2023hm} incorporate both camera and LiDAR data across multiple agents. However, HM-ViT presumes each vehicle is limited to a single sensor type, neglecting the complexities of multi-modal, multi-agent feature interactions and the effects of real-world noise. To address this gap, our AgentAlign focuses on integrating multi-agent multi-modal features and mitigating misalignment issues across heterogeneous agents to improve cooperative perception.

\section{Methodology}



\subsection{Overall Architecture\label{overall}}

The overall framework of AgentAlign is outlined in Fig.\ref{fig:framework}. Initially, agent-to-agent communication facilitates information sharing across different agents. During training, the proposed framework aligns each agent’s multi-modal features through our CFAS and HAFA module, adapts to various real-world noise conditions, and then performs the multi-agent fusion. In the testing phase, we validated our method using a real-world multi-agent dataset, further assessing the framework’s robustness by systematically testing each decomposed noise type based on the simulated V2XSet-Noise dataset.

\subsection{Multi-Modal Feature Extraction \label{Multi-modal}}
\begin{figure*}
\centering
\includegraphics[width=0.88\textwidth]{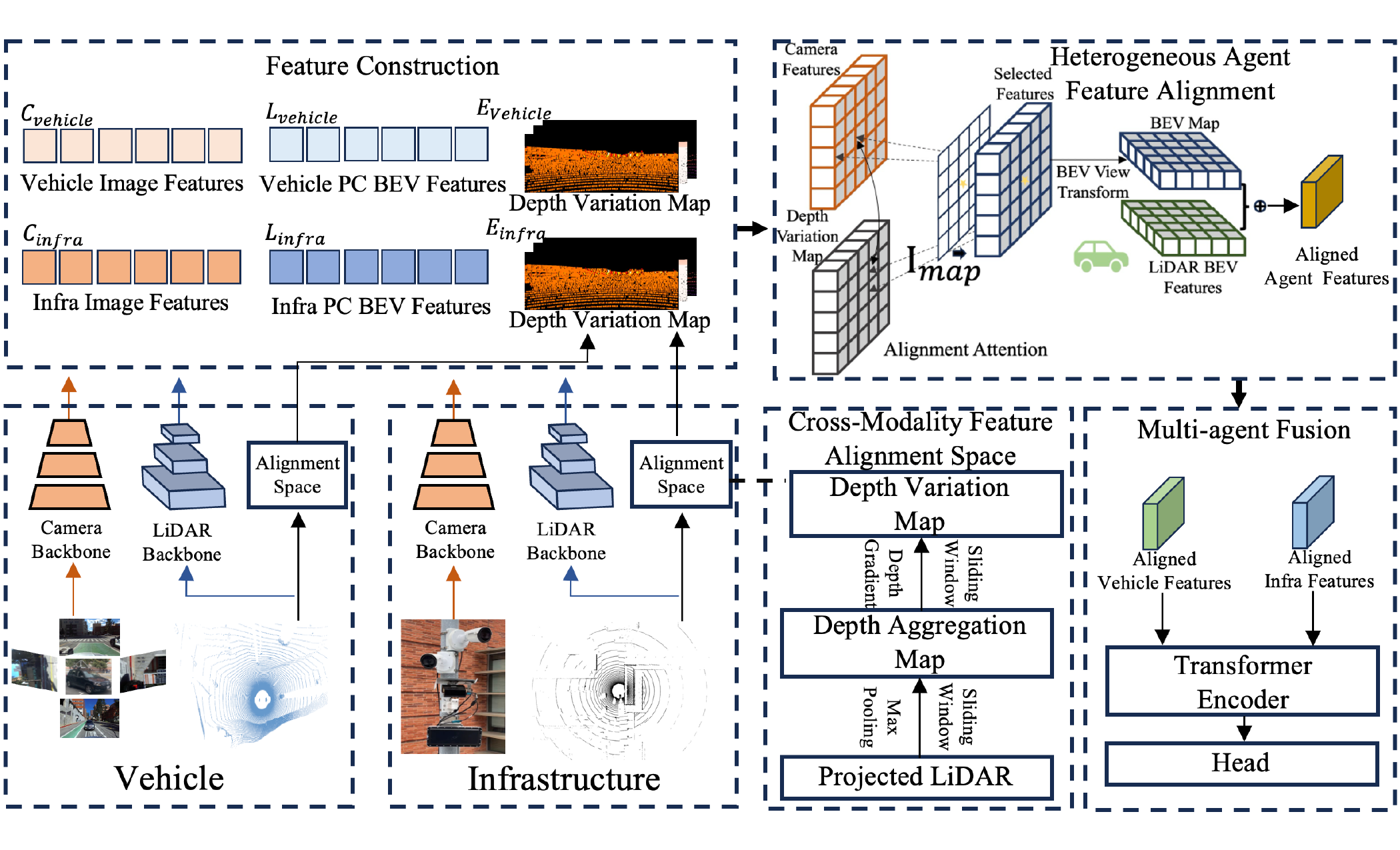}
  \caption{{\bf The detailed structure of our proposed cooperative perception network.} The proposed AgentAlign consists of feature construction, cross-modality feature alignment space, heterogeneous agent feature alignment, and multi-agent fusion. Initially, multi-modality features are extracted from both camera and LiDAR data, followed by constructing the depth variation map within CFAS. Subsequently, the HAFA mechanism dynamically selects and aligns relevant features from both the camera output and the LiDAR depth variation map. The generated features are then transformed into a bird’s-eye-view (BEV) perspective and concatenated with LiDAR features, creating an aligned feature set for each agent. Finally, each agent’s aligned feature is processed by a transformer encoder and forwarded to the detection head, which produces the 3D bounding boxes.}
\label{fig:network}
\end{figure*}

During the early stage of collaboration, every agent
within the communication networks shares metadata such as agent poses and sensor extrinsic. Then each agent starts to extract the features as shown in the camera and LiDAR backbone and feature construction module in Fig. \ref{fig:network}. We utilize the anchor-based PointPillar~\cite{lang2019pointpillars} method to extract visual features from point clouds due to its low inference latency and optimized memory usage. The raw point clouds are converted into a stacked pillar tensor, then scattered into a 2D pseudo-image, and processed by the PointPillar backbone. The backbone extracts feature information-attention maps and directly projects the LiDAR feature into bird’s-eye-view (BEV) space feature $L_i \in \mathbb{R}^{H, W, C}$ denoting agent $i$’s feature with height $H$, width $W$, and channels $C$.

The camera backbone is designed to encode input images into deep features with rich semantic information. It includes a 2D backbone for basic feature extraction and a neck module for scale-variant object representation. We employ Efficient-Net~\cite{tan2019efficientnet} for more efficient representative feature extraction. A Feature Pyramid Network (FPN) is then applied on top of the backbone to exploit multi-scale camera features and used to align the height and width of extracted camera features $C_i \in \mathbb{R}^{H, W, C_{camera}}$.

\subsection{Heterogeneous Agent Multi-Modality Alignment   \label{Alignment space}}
After extracting the LiDAR features $L_{i}$ and camera features $C_{i}$, we propose the cross-modality feature alignment space (CFAS) to align the projected LiDAR point clouds and 2D image features and improve the heterogeneous agent multi-sensor correlations. In real-world scenarios, aligning both 2D and 3D information could be difficult due to the sparsity of point clouds, the various agent-specific noise sources, and the different representation spaces of the sensor measurements. Therefore, we build this CFAS for 3D sparse point clouds, as illustrated in the cross-modality feature alignment space module in \cref{fig:network}, thereby integrating multi-sensor information to enrich the feature representations for subsequent learning tasks. 

The first stage of the proposed CFAS involves the projection of LiDAR point cloud data onto multi-view depth map representations.
This projection serves as the foundation for integrating LiDAR and multi-view images, thereby enabling richer scene understanding through the fusion of complementary sensor modalities. The projection of the LiDAR point clouds can be formulated as below:
\begin{equation}
\mathbf{p}_{\text{img}} = \mathbf{K} \cdot \mathbf{T}_{\text{lidar2img}} \cdot \mathbf{p}_{\text{lidar}}
  \label{eq:important}
\end{equation}, where $\mathbf{p}_{\text{lidar}}$ represents the LiDAR point in homogeneous coordinates. $\mathbf{T}_{\text{lidar2img}}$ is the transformation matrix mapping the LiDAR coordinates to the camera frame. $\mathbf{K}$ is the intrinsic camera matrix to provide the projection from the camera frame to the image plane. 
However, the projected LiDAR point clouds are very sparse, making it challenging for the network to learn the interaction between LiDAR and camera features. To address this issue, we densify the projected depth information through the maximum pooling operation, making the resulting depth map more dense and reducing the negative impact of having large empty areas. Specifically, a $7\times7$ sliding window is used around each pixel. For each window, the maximum depth value is computed. This process effectively fills the entire window with the maximum value found in that neighborhood, propagating depth information to nearby empty pixels. The formula to obtain the depth aggregation map $D'$ could be shown as:
\begin{equation}
D'_{\text{max}}(i, j) = \max_{(m, n) \in \Omega_{7 \times 7}} D(i + m, j + n)
  \label{eq:important}
\end{equation}
, where $\Omega_{7 \times 7}$ represents the 7x7 neighborhood around pixel $(i,j)$ and 
$D(i,j)$ is the depth map value.

To better capture the spatial variation of depth, we calculate depth gradients in four directions (up, down, left, right). This is done by applying an affine transformation to shift the depth aggregation maps, followed by computing the difference between the original and shifted depth. The gradients are then masked to remove invalid values, and the maximum absolute gradient is calculated for each direction. These gradients are subsequently concatenated with the depth aggregation map, resulting in more comprehensive representations that include both the depth information and its spatial changes. To obtain the depth variation map $G$, the depth gradient in each direction can be computed as:

\begin{equation}
g_{i,j}^{t} = d'_{i + s_i^t, j + s_j^t} - d'_{i,j}, \quad t \in \{1, 2, 3, 4\}
  \label{eq:important}
\end{equation}
, where \( d'_{i,j} \) is the \( i \)-th row and \( j \)-th column of the depth aggregation map \( D' \), and \( g_{i,j}^t \) represents the gradient of the depth aggregation map in one of the four directions (up, down, left, right).

After obtaining the depth variation map $G$, the HAFA module, as shown in \cref{fig:network}, is proposed to effectively fuse multi-modal features derived from the camera and depth variation maps, facilitating a more informative and adaptive integration of sensor data. 
We perform the fusion process by using a learned information-attention map $I_{map}$ to dynamically balance the contributions from both the camera features and the LiDAR depth variation map, enabling the network to adaptively decide which features are more relevant for a given context. Specifically, we learn this information-attention map $I_{map}$ by first concatenating the camera image features and the depth variation map along the channel dimension. Then, the convolutional operation, followed by the activation function is applied to generate the information-attention map to determine the balance between the two input features. The core principle of HAFA can be viewed as dynamically constructing a weighted feature graph where nodes represent feature elements derived from both camera and LiDAR modalities, and edges represent the attention weights assigned to different modalities. In this context, each modality (camera and LiDAR) produces its own set of feature nodes, which are then connected by the information-attention map $I_{map}$ that represents either consistent or inconsistent relationships, depending on the feature alignment across modalities. If misalignment is detected—whether due to spatial offsets, temporal lag, or perspective discrepancies—the information-attention map $I_{map}$ corrects the errors by paying more attention to reliable modalities to construct the final representation. By adaptively adjusting these edge weights through a learnable gating strategy, the network can generalize across different types of misalignment errors, dynamically selecting optimal paths through the feature graph that provides the most consistent and informative view of the environment. Finally, each heterogeneous agent contributes to aligning multi-sensor features, thereby enhancing the interactions and correlations between sensors in a multi-agent system.

With the aligned camera and LiDAR features, we design a BEV view projector module to transform the semantic pixel information into 3D space. Specifically, we use a depth estimation module to predict the depth information. We leverage a deep neural network (DNN) to explicitly predict the discrete depth distribution for each pixel by following methods like LSS \cite{philion2020lift}. Each feature pixel is then scattered into discrete points along the camera ray and rescaled according to its respective depth probabilities. After obtaining the depth estimation, we use the BEV pooling operation to aggregate all multi-view features within each BEV grid and flatten the features along the z-axis. The obtained BEV map is concatenated with the LiDAR BEV features to be used as the aligned agent features.

\subsection{Multi-Agent Feature Fusion \label{multi agent feature fusion}}
After obtaining the aligned agent features, the goal of the multi-agent fusion module is to further integrate the multi-source multi-modal BEV features to enhance the perception of the surroundings. We utilize a self-attention mechanism as the shared network to fuse each agent’s aligned features. The self-attention module calculates correlation scores among the aligned features, constructing a local graph for each agent's feature vector within the aligned feature map. In this graph, we calculate the relationship between the aligned feature vectors at the same spatial locations across different connected agents. Such a fusion method facilitates the effective integration of the intermediate features. The detection head then processes these multi-agent fused BEV features to generate 3D bounding boxes. 

\subsection{V2X Multifactorial Noise \label{multi noise}}

To evaluate the impact of different types of noise and assess the robustness of AgentAlign, we constructed a new dataset, V2XSet-Noise, based on simulated V2XSet\cite{xu2022v2x} to explicitly account for real-world sensor noise in V2X communication scenarios. The new V2XSet-Noise will also be released on our paper website. 

\textbf{Calibration Error}
The sensor calibration error \cite{an2022online} comes from the extrinsic metrics and it can cause misaligned camera images and LiDAR point clouds. In reality, factors such as shocks, vibrations, mechanical fatigue, or loosening from prolonged vehicle operation, as well as manufacturing assembly tolerances and sensor installation inaccuracies, inevitably lead to calibration errors between sensors. This problem can be particularly series on the infrastructure side due to the high cost of continuous maintenance. Such errors can occur in all six dimensions: x,y,z, pitch, roll, and yaw. To simulate these calibration errors and follow the canonical initial errors \cite{calib_2000,koide2023general,an2022online}, we introduce random perturbations ranging from -0.5 to 0.5 degrees for pitch, roll, and yaw angle and add the -0.5m to 0.5m to the x, y, and z.

\textbf{Wind-induced Vibration} Infrastructure sensors are often susceptible to wind-induced disturbances, particularly those positioned at elevated locations. To simulate this wind-induced vibration, we follow the methodology outlined by \cite{chen2005wind, ZHANG2022673}, introducing sinusoidal signals to the sensors. Specifically, we apply a sequential addition of sinusoidal perturbations to both the LiDAR and cameras respectively at a frequency of 2 Hz, replicating the effects of wind-induced vibrations experienced by high-mounted sensors. 

\textbf{Infrastructure Camera Distortion}
Infrastructure cameras are often prone to perspective and optical distortions \cite{valente2015perspective}, particularly when mounted at elevated or angled positions. To simulate real-world distortions in infrastructure cameras, we model the camera using the pinhole model, adjusting the homography to alter the camera’s perspective effect. Additionally, we simulate optical distortions by modifying radial and tangential distortion parameters. Such a method adjusts the width of the image while keeping the height unchanged, simulating the effect of oblique viewing angles.

\textbf{Vehicle LiDAR Distortion}
Vehicle-mounted LiDAR sensors can experience motion-based distortions, particularly when the vehicle is accelerating, braking, or turning. To replicate these distortions, we introduce a segmented motion compensation method for mechanical LiDARs, dividing the point cloud data into multiple regions. Each region undergoes a unique transformation based on linear and angular perturbations, mimicking the effects of vehicle dynamics on the LiDAR point cloud \cite{9811273}. 

\textbf{Time Sychronization}
In V2X scenarios, the heterogeneity of sensors introduces certain triggering errors during software synchronization. It is common for camera and LiDAR sensors to be triggered at different times \cite{dair-v2x, cooperfuse}. To replicate this, we simulate asynchronous triggering for both infrastructure and vehicle sensor suits. In our simulation, the maximum trigger delay between the LiDAR and camera sensors is set to 0.1 seconds, reflecting realistic temporal discrepancies.

\textbf{Systematic Error}
In real V2X systems, sensors may experience fixed misalignment errors due to external environmental interference \cite{zhang2016systematic}. For example, sensors may shift to an incorrect position due to system external forces such as temperature change, moisture, and corrosion, resulting in a persistent offset that affects detection accuracy. Because such errors are typically bounded and result from random environmental effects, we apply fixed translation and rotation errors with noise uniformly selected between -0.1 and 0.1 meters and -0.1 and 0.1 degrees. This approach captures the equal likelihood of shifts in any direction within a realistic range, avoiding directional bias and ensuring that noise remains within physically plausible limits.


\section{Experiments}

\subsection{Implementation Details}
We implement our module with Efficient-Net as the 2D backbone for the multi-view image encoder and PointPillars \cite{lang2019pointpillars} as the LiDAR feature extraction backbone. We use 1664 $\times$ 960 as the image size and the voxel size following the default settings in \cite{lang2019pointpillars, xu2021opv2v}. 
We first train the LiDAR stream and camera stream with multi-view image input and LiDAR point clouds input, respectively. We then train the HAFA to align both images and LiDAR features for another 10 epochs that inherit the weights from the trained two streams. This staged training approach aims to reduce computational complexity when multiple agents and their corresponding data are required to be involved concurrently.

\subsection{Evaluation and Dataset}
The performance of AgentAlign is evaluated on both the real-world multi-agent V2X-Real, the simulated V2XSet-Noise, and the noisy OPV2V dataset. The evaluation is conducted within the range of -100m to 100m in the x direction and -40m to 40m in the y direction of the ego coordinate frame. Following the V2X-Real benchmark evaluation protocol, we adopt IoU thresholds (IoU=0.3, 0.5) for the real-world AP calculation to reflect the model’s perception capability. 



Our V2XSet-Noise dataset is designed to simulate various types of noise encountered in real-world scenarios. We build this dataset on top of the simulated V2XSet, where two to seven intelligent agents can communicate in each sequence. We corrupt the V2XSet data by following the method outlined in subsection \ref{multi noise}. We add wind-induced vibration and systematic error to the infrastructure's LiDAR and motion distortion to the vehicle's LiDAR. Distorted camera images are generated for the infrastructure. Additionally, we corrupt the extrinsic matrices for both vehicles and infrastructure. Time asynchronization is introduced to both vehicle and infrastructure cameras and LiDAR data. All noise is added in a controlled manner, resulting in a total of 11,447 frames in our dataset. Similarly, noise corruption is applied at the same levels to the OPV2V dataset to ensure a fair comparison with existing methods.


\subsection{Comparison with State-of-the-art methods}
Our proposed method was tested on the diverse scenes of the datasets, with quantitative performance discussed in this subsection. Tables~\ref{table:v2xreal}, \ref{table:v2vsim}, and \ref{table:v2xsim} provide a comprehensive evaluation of the proposed method compared to state-of-the-art (SOTA) approaches across different datasets and scenarios. Table ~\ref{table:v2xreal} compares the performance on the V2X-Real dataset under real-world multifactorial noise using different fusion strategies. The proposed method achieves the highest performance with 62.6\% at AP@0.3 and 60.4\% at AP@0.5, surpassing the existing SOTA V2X-Real approach. These results on the noisy real-world dataset demonstrate the robustness of the proposed HAFA strategy, especially in dealing with multifactorial noise conditions, highlighting the effectiveness of our AgentAlign in real-world applications. 

In Table~\ref{table:v2vsim},  we compare the current SOTA multi-agent perception methods by incorporating vehicle LiDAR distortion and infrastructure LiDAR vibration errors.       
In the noisy V2V dataset, the proposed AgentAlign achieves the highest AP values, with 89.9\% at AP@0.5 and 66.3\% at AP@0.7, significantly outperforming other methods. On the noisy V2X dataset, shown in Table~\ref{table:v2xsim}, the proposed method achieves an AP@0.5 of 82.4\% and an AP@0.7 of 60.4\%, surpassing all other methods. The proposed AgentAlign consistently outperforms others across all datasets and metrics, demonstrating more resilient inter-agent sensor correlations and effectiveness of the proposed CFAS and HAFA in both real-world noise conditions and simulated environments. 









\begin{table}[]
\caption{\textbf{Performance comparison on the V2X-Real dataset with real-world multifactorial noise.} {We report the numbers (\%) in AP under different levels.}}
\begin{tabular}{l|l|ll}
\rowcolor[HTML]{EFEFEF}

 \hline
Cam Method     & \hspace{5 mm}Fusion & AP@0.3 & AP@0.5 \\ \hline

Point Pillar\cite{lang2019pointpillars}  & \hspace{2 mm}No Fusion    & 54.0   & 49.8   \\ 
OPV2V\cite{xu2021opv2v}  & \hspace{1 mm}Late Fusion    & 47.4   & 44.4   \\
F-Cooper\cite{fcooper}  & \hspace{1 mm}Intermediate    & 42.7   & 40.3   \\
V2X-Real\cite{xiang2024v2x} & \hspace{1 mm}Intermediate    & 58.6   & 55.3   \\
\hline
AgentAlign(Ours)  & \hspace{1 mm}Intermediate   & 62.6   & 60.4   \\

\hline

\end{tabular}
\label{table:v2xreal}

\end{table}








\begin{table}[]
\caption{\textbf{Performance comparison on the noisy V2V simulation dataset.} {We report the numbers (\%) in AP under different levels.}}
\begin{tabular}{l|l|ll}
\rowcolor[HTML]{EFEFEF}

\hline
Methods     & \hspace{5 mm}Fusion & AP@0.5 & AP@0.7 \\ \hline

Point-Pillar\cite{lang2019pointpillars}  & \hspace{1.5 mm}Late Fusion   & 54.4  &  28.3   \\  
OPV2V\cite{xu2021opv2v}  & \hspace{1 mm}Intermediate   & 79.3   & 51.9   \\ 

F-Cooper\cite{fcooper}  & \hspace{1 mm}Intermediate    & 81.7   & 58.2   \\

Where2Comm\cite{hu2022where2comm}  & \hspace{1 mm}Intermediate   & 79.2   & 46.8   \\

\hline

AgentAlign(Ours)  & \hspace{1 mm}Intermediate    & 89.9   & 66.3   \\
\hline

\end{tabular}
\label{table:v2vsim}

\end{table}





\begin{table}[]
\caption{\textbf{Performance comparison on the V2XSet-Noise simulation dataset.} {We report the numbers (\%) in AP under different levels.}}
\begin{tabular}{l|l|ll}
\rowcolor[HTML]{EFEFEF}

\hline
Methods     & \hspace{5 mm} Fusion & AP@0.5 & AP@0.7 \\ \hline
PointPillar \cite{lang2019pointpillars}  & \hspace{1.5 mm}Late Fusion    & 51.4   & 26.7   \\
OPV2V \cite{xu2021opv2v} & \hspace{1 mm}Intermediate    & 77.1   & 52.1  \\
F-Cooper\cite{fcooper}  & \hspace{1 mm}Intermediate    & 77.5   & 52.7   \\
Where2Comm\cite{hu2022where2comm}  & \hspace{1 mm}Intermediate    & 74.6   & 47.5   \\
\hline
AgentAlign(Ours)  & \hspace{1 mm}Intermediate     &82.4    &  60.4  \\
\hline

\end{tabular}
\label{table:v2xsim}

\end{table}

\begin{table}[]
\caption{\textbf{The sub-module analysis of our proposed model.} {We verify the effectiveness of each sub-module in V2X-Real dataset.}}
\scalebox{0.889}{%

\begin{tabular}{llll|ll}
\rowcolor[HTML]{EFEFEF}
\hline
\multicolumn{4}{l|}{\hspace{20 mm}Sub-module}   & \multicolumn{2}{l}{Average Precision} \\ \hline
\rowcolor[HTML]{EFEFEF}
Cam. & Depth Agg. &Depth Var. &HAFA & AP@0.3               & AP@0.5               \\ \hline
\hspace{3 mm}\checkmark     &        &             &   & 39.0              & 27.2              \\

\hspace{3 mm}\checkmark   &    \hspace{4 mm}\checkmark    &           &     & 45.6              & 34.0              \\

\hspace{3 mm}\checkmark   &    \hspace{4 mm}\checkmark    &     \hspace{6 mm}\checkmark      &     & 47.7              & 36.1              \\

\hspace{3 mm}\checkmark   &    \hspace{4 mm}\checkmark    &     \hspace{6 mm}\checkmark      &  \hspace{3 mm}\checkmark   & 49.9              & 38.4              \\
\hline
\end{tabular}%
}
\label{table:ablation}

\end{table}






\subsection{Abalation Studies}

\renewcommand{\arraystretch}{1}

\begin{figure}[t]
\centering
\includegraphics[width=0.48\textwidth]{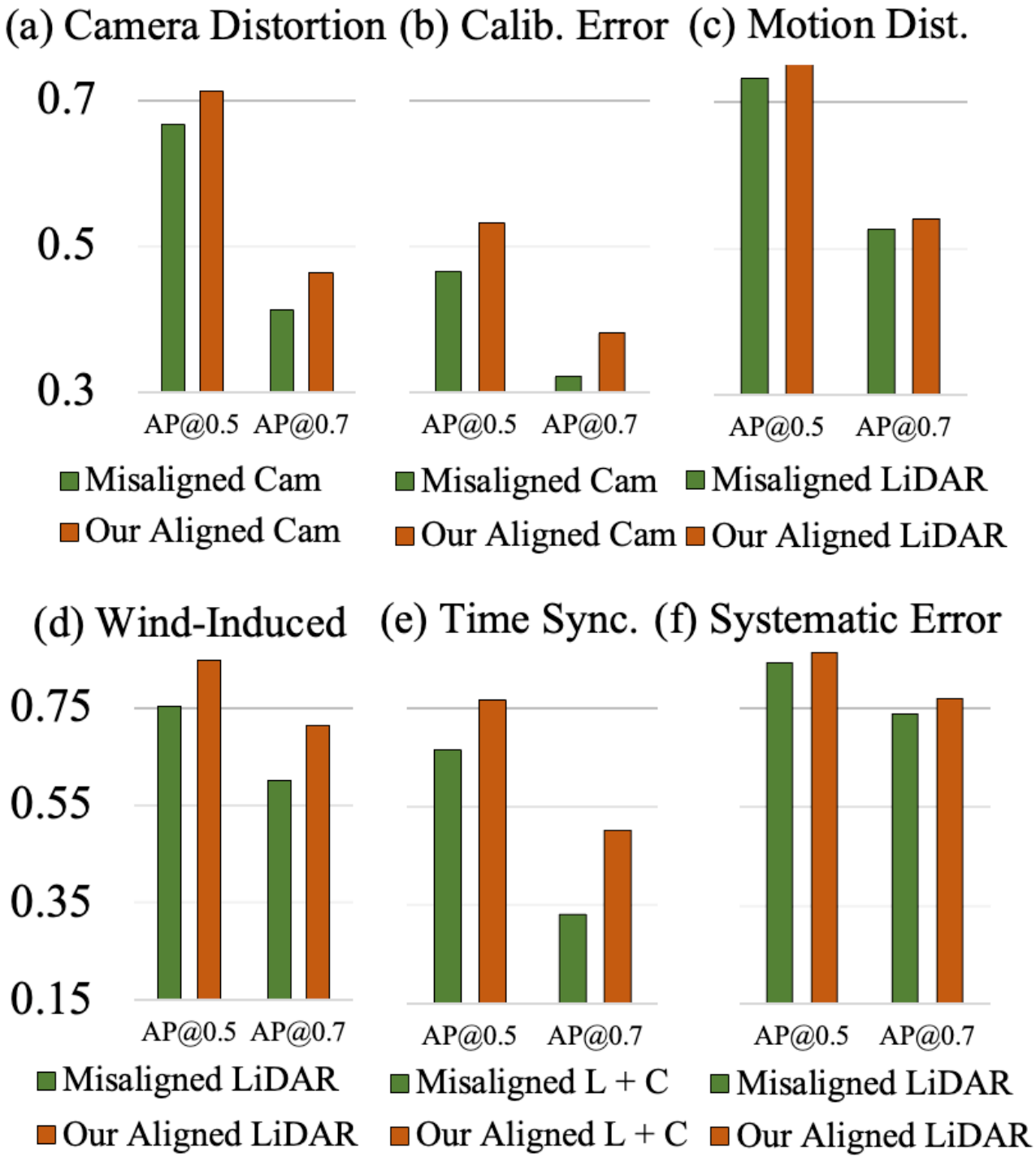}
  \caption{{\bf The decomposed noise study.} We analyze the impact of each decomposed noise by comparing the cooperative perception performance using our aligned features against the misaligned features in the V2XSet-Noise dataset.}
\label{fig:error}
\end{figure}

To further investigate the effectiveness of each sub-module of our proposed AgentAlign, we analyze the performance on the V2X-Real dataset as shown in Table \ref{table:ablation}. The sub-modules include pure camera input, depth aggregation map, depth variation map, and HAFA. When only the camera is used, the model achieves 39.0\% and 27.2\% at AP@0.3 and AP@0.5. Adding the dense aggregation map module increases the AP@0.3 to 45.6\%. Incorporating both the depth aggregation (Depth Agg.) map and depth variation (Depth Var.) map further improves 2.1\% by AP@0.3. When all sub-modules, including the HAFA mechanism, are utilized, the model achieves the highest performance, with 49.9\% and 38.4\% at AP@0.3 and AP@0.5. This analysis highlights the contribution of each sub-module, further demonstrating the significant performance improvements of our proposed CFAS and HAFA methods and their role in achieving more compact multi-sensor correlations under multifactorial noise and misalignment scenarios.

\subsection{V2X Multifactorial Noise Studies}
In this section, we analyze the impact of each decomposed noise. \cref{fig:error}(a) shows the performance comparison on the V2X simulation noisy test dataset, where camera perspective distortion is introduced. When the camera features are not aligned, the camera-branch method only achieves an AP@0.5 of 66.8 \% and an AP@0.7 of 41.3 \%. In contrast, the proposed method, aligning the camera features, significantly outperforms the baseline, achieving 71.3 \% at AP@0.5 and 46.4 \% at AP@0.7.
\cref{fig:error}(b) presents the performance comparison with the simulated extrinsic error between the LiDAR and camera. With the extrinsic error, the camera branch records 46.7\% at AP@0.5. By incorporating the proposed alignment method, our aligned camera features lead to an improvement of 6.5\% at AP@0.5. These results demonstrate the effectiveness of correcting camera-related misalignment errors, emphasizing that accurate alignment between LiDAR and camera plays a vital role in achieving high detection precision and reducing the impact of sensor misalignment.
Similarly, in \cref{fig:error}(c), (d), (e), and (f), our method also outperforms the SOTA LiDAR multi-agent perception in the motion distortion, wind-induced vibration, time synchronization, and systematic errors. The results presented in each of the noisy settings clearly indicate that the proposed AgentAlign lead to consistent improvements across all noisy scenarios and the generalization capability of our CFAS and HAFA. These findings highlight that AgentAlign enhances the consistency of interactions and correlations between heterogeneous agent sensors.


\section{Conclusion}
In this work, we introduce AgentAlign, a robust multi-modality multi-agent cooperative perception framework, designed to address agent-specific misalignment issues and multifactorial noise in V2X scenarios. Our AgentAlign leverages a cross-modality feature alignment space (CFAS) and a heterogeneous agent feature alignment(HAFA) mechanism to dynamically integrate and harmonize 3D LiDAR and 2D camera features. To better evaluate the generalization capability of AgentAlign, we constructed a novel, controllable V2XSet-Noise dataset that simulates various realistic sensor imperfections to systematically assess AgentAlign under diverse environmental conditions. Extensive experiments conducted on both real-world V2X-Real and simulated V2XSet-Noise datasets demonstrate the effectiveness of our proposed framework, achieving state-of-the-art performance in challenging cooperative detection tasks. 

{
}

\clearpage
\setcounter{page}{1}
\setcounter{section}{0} 
\setcounter{equation}{0} 
\maketitlesupplementary

\section{Sensor Calibration Error}
\label{sec:calibration_error}

Accurate sensor calibration is essential for aligning data from different sensors, such as cameras and LiDARs, into a common coordinate frame. Miscalibration in the extrinsic parameters can lead to misaligned sensor data, negatively impacting the performance of perception algorithms. Generally, sensors undergo calibration before operation. However, during prolonged periods of operation, calibration errors may arise from various factors, including mechanical vibrations, environmental conditions, or installation inaccuracies. For infrastructure sensors, maintaining accurate calibration metrics is particularly challenging due to high maintenance costs. Calibration errors can affect all six degrees of freedom: translations along the \(x\), \(y\), and \(z\) axes, and rotations around the pitch $\theta$, roll $\phi$, and yaw $\psi$ angles. In this section, we only discuss the calibration errors between cameras and LiDARs.

Denote $\mathbf{T}_l^{c}$ as the homogeneous transformation matrice that maps points from the LiDAR frame to the camera frame, defined as:
\begin{equation}
    \label{tcl}
        \mathbf{T}_l^{c} = 
    \begin{bmatrix}
        \mathbf{R}_l^{c} & \mathbf{t}_l^{c} \\
        \mathbf{0} & 1
    \end{bmatrix},
\end{equation}
where,
\begin{itemize}
    \item \( \mathbf{R}_l^{c} \in \mathbb{R}^{3 \times 3} \) is the rotation matrix from LiDAR to camera frame.
    \item $\mathbf{t}_l^{c} = [t_x, t_y, t_z]^\top \in \mathbb{R}^{3 \times 1} $ is the translation vector that represents the displacement between the camera and LiDAR frames.
    \item $\mathbf{0} \in \mathbb{R}^3$ that ensures compatibility with homogeneous coordinates.  
\end{itemize}

To transform points from  LiDAR coordinates \( \mathbf{X}_l \in \mathbb{R}^3 \) to camera coordinates \( \mathbf{X}_c \in \mathbb{R}^3 \), we use the following homogeneous transformation formula:

\begin{align}
    \widetilde{\mathbf{X}_c} &= \mathbf{T}_l^{c} \cdot \widetilde{\mathbf{X}_l},
\end{align}
where, \(\widetilde{\mathbf{X}_c}, \widetilde{\mathbf{X}_l} \in \mathbb{R}^4 \) represent the camera and LiDAR points, respectively, in homogeneous coordinates.



To emulate calibration errors between cameras and LiDARs, we perturb both the rotation matrix \( \mathbf{R} \) and the translation vector \( \mathbf{t} \) in the transformation matrix \( \mathbf{T}_l^{c} \). We use uniform distribution $\mathcal{U}$ to consider equal likelihood for all potential errors within a specified range, reflecting an unbiased representation of calibration inaccuracies and providing a realistic model for random perturbations. The uniformly sampled noise is applied to the rotation angles for roll (\( \xi \)), pitch (\( \theta \)), and yaw (\( \psi \)) to generate the random offset noise:
    \begin{align*}
        \xi' &= \xi + \Delta {\xi}, & \Delta {\xi} &\sim \mathcal{U}(-0.5^\circ, 0.5^\circ) \\
        \theta' &= \theta + \Delta {\theta}, & \Delta {\theta} &\sim \mathcal{U}(-0.5^\circ, 0.5^\circ) \\
        \psi' &= \psi + \Delta {\psi}, & \Delta {\psi} &\sim \mathcal{U}(-0.5^\circ, 0.5^\circ)
    \end{align*}
    
The rotation matrix \( \mathbf{R'} \) is recalculated using the perturbed angles:
    \[
    \mathbf{R'} = R_{\psi}' \cdot R_{\theta}' \cdot R_{\xi}'
    \]
    where:
    \begin{align*}
        R_{\xi}' = \begin{bmatrix}
        1 & 0 & 0 \\
        0 & \cos\xi' & -\sin\xi' \\
        0 & \sin\xi' & \cos\xi'
        \end{bmatrix} \\
        R_{\theta}' = \begin{bmatrix}
        \cos\theta' & 0 & \sin\theta' \\
        0 & 1 & 0 \\
        -\sin\theta' & 0 & \cos\theta'
        \end{bmatrix} \\
        R_{\psi}' = \begin{bmatrix}
        \cos\psi' & -\sin\psi' & 0 \\
        \sin\psi' & \cos\psi' & 0 \\
        0 & 0 & 1
        \end{bmatrix}
    \end{align*}
    
where $R_{\psi}'$ is the rotation matrix around the $Z$-axis(yaw), $R_{\theta}'$ is the rotation matrix around the $Y$-axis(pitch), $R_{\xi}'$ is the rotation matrix around the $X$-axis(roll).

Uniformly randomly sampled offsets are also introduced to the translation vector $\mathbf{t}_c^{l'}=[t_x', t_y', t_z']^\top$ by:
\begin{align*}
    t_x' &= t_x + \Delta {t_x}, & \Delta {t_x} &\sim \mathcal{U}(-0.5\,\text{m}, 0.5\,\text{m}) \\
    t_y' &= t_y + \Delta {t_y}, & \Delta {t_y} &\sim \mathcal{U}(-0.5\,\text{m}, 0.5\,\text{m}) \\
    t_z' &= t_z + \Delta {t_z}, & \Delta {t_z} &\sim \mathcal{U}(-0.5\,\text{m}, 0.5\,\text{m})
\end{align*}

After noise is added, we get the new perturbed transformation matrix $\mathbf{T}_l^{c'}$ by Equ.~\ref{tcl}. We transform the LiDAR coordinates \( \mathbf{X}_l \) into the perturbed camera coordinates \( \mathbf{X}_c' \):
\begin{equation}
    \mathbf{X}_c' = \mathbf{T}_l^{c'} \cdot \mathbf{X}_l,
\end{equation}

Therefore, we can emulate the sensor calibration errors by modifying $\mathbf{T}_l^c$. 











\section{Wind-Induced Vibration Noise}
\label{sec:wind_noise}
\begin{figure*}
\centering
\includegraphics[width=0.88\textwidth]{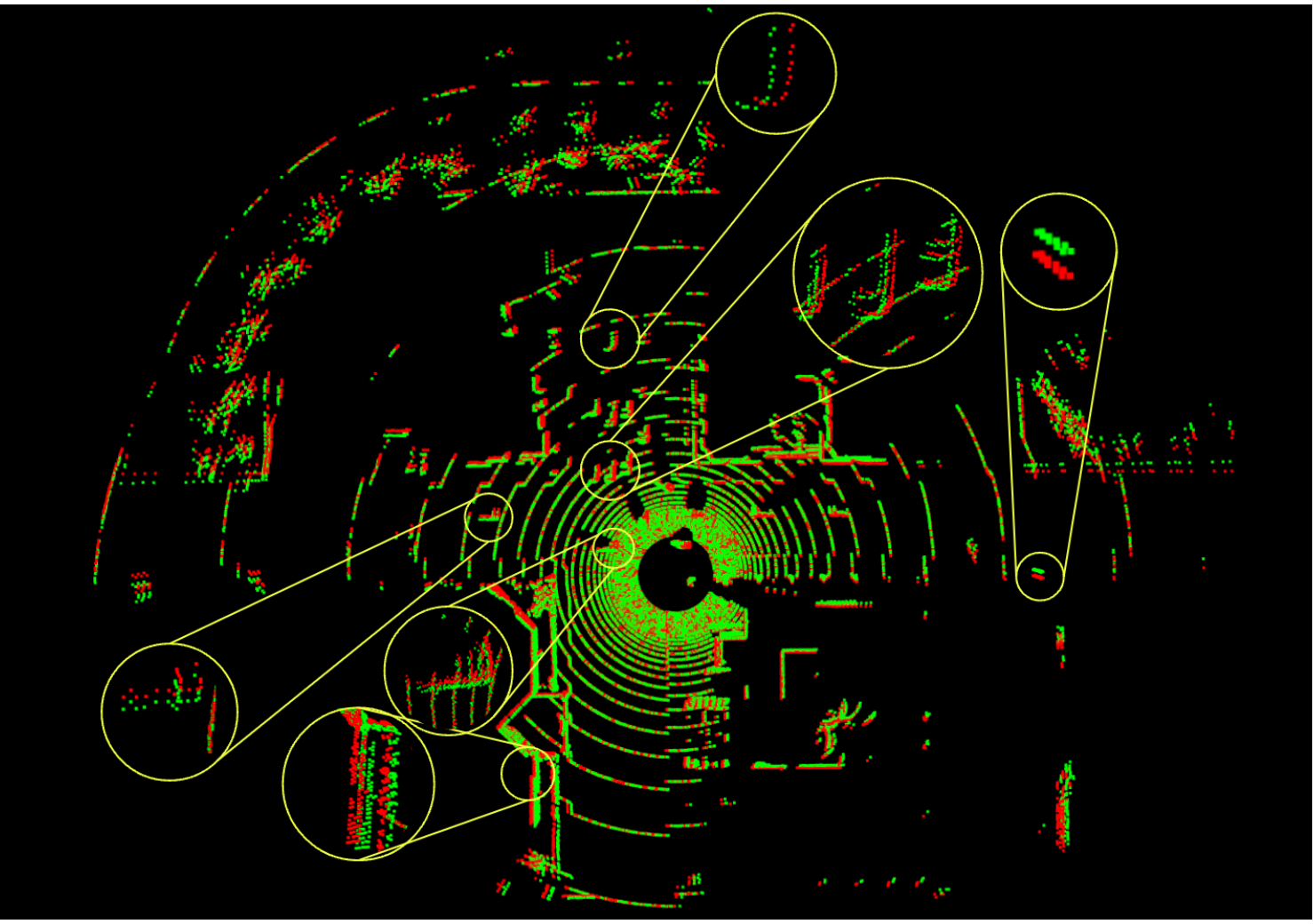}
\caption{
{\bf Visual representation of Wind-introduced vibration} The red points indicate the original point cloud data, while the green points visualize the data after applying wind-introduced vibration. In the figure, we magnify specific regions to clearly illustrate the impact of vibration on the sensor data.
}
\label{fig:Wind_distortion}
\end{figure*}

Infrastructure sensors mounted at elevated positions are susceptible to wind-induced vibrations, affecting the accuracy of the collected data. To emulate these real-world impacts on LiDAR and camera sensors, we introduce rotational and positional noise into our multi-agent system. 

Similar to Equ.\ref{tcl}, we denote $\mathbf{T}$ as a homogeneous transformation matrix that maps points in the world coordinate ($\mathbf{X}_w$) to the LiDAR coordinate($\mathbf{X}_l$).


We calculate the wind-induced perturbed rotation matrix \( R' \) by calculating sinusoidal oscillations in the roll (\( \xi \)), pitch (\( \theta \)), and yaw (\( \psi \))  angles:
\begin{align}
    \label{w1}
    \xi^{'}(t) &= A \cdot \sin(2\pi f t + \Delta \xi) + \xi,\\
    \label{w2}
    \theta^{'}(t) &= A \cdot \sin(2\pi f t + \Delta \theta) + \theta, \\
    \label{w3}
    \psi^{'}(t) &= A \cdot \sin(2\pi f t + \Delta \psi) + \psi, 
\end{align}

where, $A = 0.5^\circ $ is the amplitude of the rotational vibration. \( t \) is a time variable, meaning that the noise added to $\psi, \theta, \xi$ are time-variant. f is the frequency of the vibration.
$\Delta \psi, \Delta \theta, \Delta \xi$ are random phase shifts uniformly sampled from \([0, 2\pi]\).

White Gaussian noise is added to the point cloud translation to replicate the random micro-movements caused by sensor vibrations. These perturbations mimic the unpredictable, fine-grained shifts that occur during real-world vibration scenarios, enhancing the realism of the simulated data and providing a more robust test environment for perception algorithms. The perturbed translation vector \( \mathbf{t}' \) incorporates Gaussian noise:
\begin{align}
    \label{t1}
    \mathbf{t}_x' &= \mathbf{t}_x + \Delta \mathbf{t}_x, & \Delta \mathbf{t}_x &\sim \mathcal{N}(0, 0.02^2) \\
    \label{t2}
    \mathbf{t}_y' &= \mathbf{t}_y + \Delta \mathbf{t}_y, & \Delta \mathbf{t}_y &\sim \mathcal{N}(0, 0.02^2) \\
    \label{t3}
    \mathbf{t}_z' &= \mathbf{t}_z + \Delta \mathbf{t}_z, & \Delta \mathbf{t}_z &\sim \mathcal{N}(0, 0.01^2)
\end{align}

Based on Equ.\ref{w1} $\sim$ \ref{t3}, we can get the new homogeneous transformation matrix $\mathbf{T}^{'}$ and wind-induced point cloud data $\mathbf{X}_{l}^{'}$ in LiDAR coordinates: 
\begin{align}
    \mathbf{T}^{'} = f(\mathbf{R}^{'}, \mathbf{t}^{'}), \\
    \mathbf{X}_{l}^{'} = f(\mathbf{T}^{'},\mathbf{X}_w)
\end{align}
To illustrate the effects of wind-induced vibrations, we include visualizations in Figure~\ref{fig:Wind_distortion}.

 To replicate the effect of vibrations on camera images, we apply pixel-wise translations in both the horizontal and vertical directions to a sequence of images. The vibration effect is modeled using sinusoidal functions, with the following steps and formulations. The displacement for each frame is computed using sinusoidal functions with specified amplitude and frequency. The horizontal and vertical shifts (\(\Delta u\) and \(\Delta v\)) are given by:
\begin{align}
    \Delta x &= A \cdot \sin(2 \pi f t + \Delta \phi_{1}), \\
    \Delta y &= A \cdot \sin\left(2 \pi f t  + \Delta \phi_{2} \right),
\end{align}
where, \(A\) is the amplitude of vibration, corresponding to 1.5\% of the imagersolution. \(f\) is the vibration frequency (in Hz). \(t = \frac{\text{frame index}}{\text{sampling rate}}\) is the time step corresponding to each frame. $\Delta \phi_{1}$ and $\Delta \phi_{2}$ are random phase uniformly sampled from \([0, 2\pi]\) to simulate irregular vibration patterns.
The computed displacements are applied to the image using the following affine transformation matrix:
\begin{equation}
    \mathbf{A}_c = 
    \begin{bmatrix}
    1 & 0 & \Delta u \\
    0 & 1 & \Delta v
    \end{bmatrix}.
\end{equation}
This matrix defines the pixel shifts in the horizontal (\(u\)) and vertical (\(v\)) directions. Each image is transformed using affine transformation with the translation matrix \(\mathbf{A}_c\) to the image, resulting in a displacement by the calculated \(\Delta u\) and \(\Delta v\). The transformation is applied frame-by-frame, with the time step \(t\) determined by the frame index and sampling rate.

\section{Infrastructure Camera Distortion}
\label{sec:rationale}
\begin{figure*}

\centering
\includegraphics[width=0.88\textwidth]{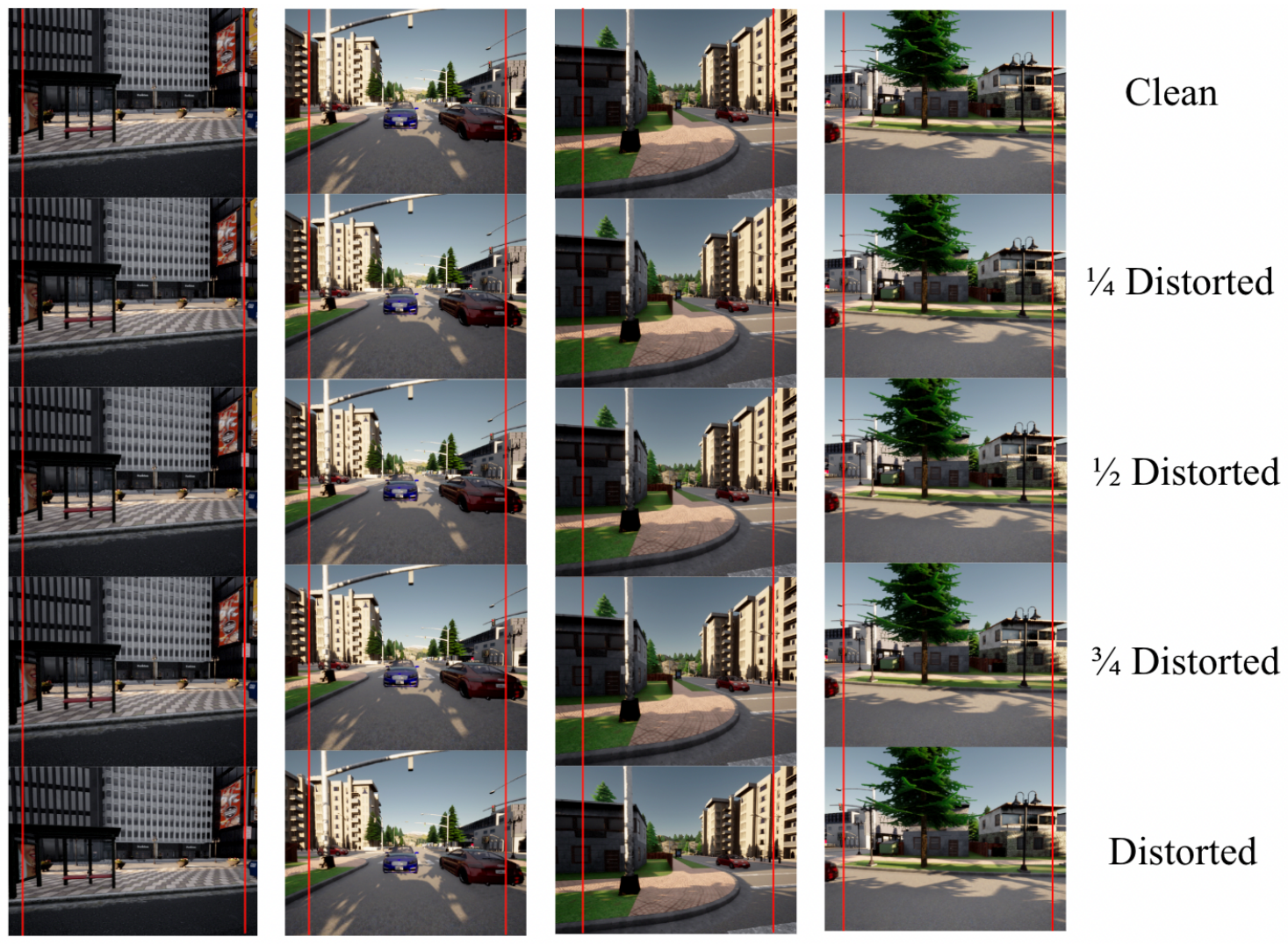}
\caption{
{\bf Visual representation of perspective distortion levels in images captured by infrastructure cameras.} The "Clean" column depicts undistorted images, while the subsequent columns demonstrate increasing levels of distortion: Minimal Distortion, Low Distortion, Moderate Distortion, and Full Distortion. The vertical red line serves as a reference to clearly illustrate the varying levels of distortion in the camera images.
}
\label{fig:camera_distortion}
\end{figure*}

Elevated or angled infrastructure cameras frequently capture images that exhibit perspective distortions due to their mounting position and viewing angle. These distortions impact the spatial accuracy of visual data, particularly in scenarios such as traffic monitoring, urban surveillance, or capturing objects from non-orthogonal perspectives. Such distortions are characterized by skewed or stretched views of objects, which can reduce the camera-based perception algorithms and reliability of downstream tasks.

As noted in Tan et al. (2017)~\cite{Tan2017}, "there is no established quantitative definition for perspective distortion." If a photograph is not taken straight-on, even with a perfect rectilinear lens, rectangles will appear as trapezoids: lines are imaged as lines, but the angles between them are not preserved. 

To address this, we define perspective distortion using our own approach. The transformation is represented based on the dimensions of the image, where \( H \) is the height and \( W \) is the width. The four corners of the original undistorted image are defined as the natural boundaries of the frame, expressed mathematically as:

\begin{equation}
\mathbf{P}_{\text{src}} = 
\begin{bmatrix}
(0, 0)^\top \\ 
(W, 0)^\top \\ 
(W, H)^\top \\ 
(0, H)^\top
\end{bmatrix}
\end{equation}

To simulate perspective distortion\cite{valente2015perspective}, the top corners are shifted horizontally outward, while the bottom corners remain fixed to emulate the effect of oblique viewing angles from elevated cameras. The destination points for the distorted image are:

\begin{equation}
\mathbf{P}_{\text{dst}} = 
\begin{bmatrix}
(-\alpha W, 0)^\top \\ 
((1 + \alpha)W, 0)^\top \\ 
(W, H)^\top \\ 
(0, H)^\top
\end{bmatrix}
\end{equation}
where \( \alpha \) is the distortion parameter that controls the outward displacement of the top corners.

The homography transformation matrix \( H \) maps each pixel from the original image to its distorted counterpart:

\begin{equation}
\begin{bmatrix}
x' \\ y' \\ 1
\end{bmatrix}
=
\mathbf{H}
\cdot
\begin{bmatrix}
x \\ y \\ 1
\end{bmatrix}
\end{equation}

\( \mathbf{H} \) can be computed using the source points \( \mathbf{P}_{\text{src}} \) and destination points \( \mathbf{P}_{\text{dst}} \) by setting up the equations for each of the four points:

\subsubsection*{Point 1: \( (x_1, y_1) = (0, 0) \rightarrow (x_1', y_1') = (-\alpha W, 0) \)}

\subsubsection*{Point 2: \( (x_2, y_2) = (W, 0) \rightarrow (x_2', y_2') = ((1+\alpha) W, 0) \)}

\subsubsection*{Point 3: \( (x_3, y_3) = (W, H) \rightarrow (x_3', y_3') = (W, H) \)}

\subsubsection*{Point 4: \( (x_4, y_4) = (0, H) \rightarrow (x_4', y_4') = (0, H) \)}

Solved the homography matrix:

\[
\mathbf{H} =
\begin{bmatrix}
1 + 2\alpha & \dfrac{\alpha W}{H} & -\alpha W \\
0 & 1 + 2\alpha & 0 \\
0 & \dfrac{2\alpha}{H} & 1
\end{bmatrix}
\]

The parameter \( \alpha \) is adjusted to achieve the following levels of distortion, which determines the horizontal displacement of the top corners.:
\begin{itemize}
    \item Minimal Distortion (1/4): \( \alpha = 0.007 \): Simulates a slight horizontal shift for top corners.
    \item Low Distortion (1/2): \( \alpha = 0.014 \): Represents a moderate horizontal shift, introducing noticeable skew.
    \item Moderate Distortion (3/4): \( \alpha = 0.021 \): Creates a stronger perspective effect, resembling steep angles.
    \item Full Distortion (1): \( \alpha = 0.028 \): Mimics extreme skewing caused by oblique camera angles.
\end{itemize}

To illustrate the effects of the perspective distortion, we include the following visualizations in Fig.\ref{fig:camera_distortion}.

\section{Vehicle LiDAR Distortion} 
\begin{figure*}
\centering
\includegraphics[width=0.88\textwidth]{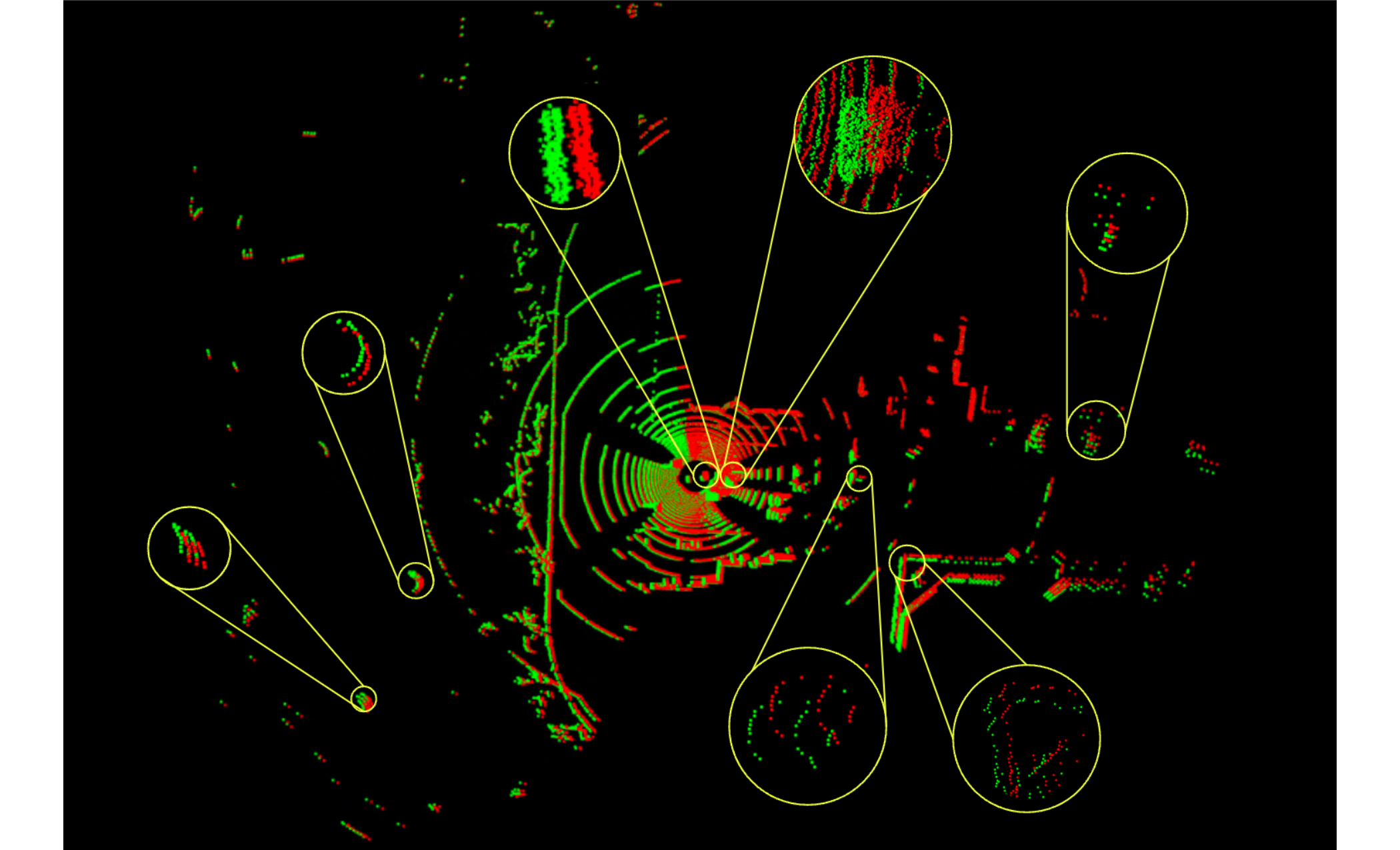}
\caption{
{\bf Visual representation of Motion-introduced distortion} The red points indicate the original point cloud data, while the green points visualize the data after applying motion-introduced vibration. In the figure, we magnify specific regions to clearly illustrate the impact of motion on the sensor data.
}
\label{fig:Motion_distortion}
\end{figure*}
Vehicle-mounted LiDAR sensors frequently experience motion-induced distortions due to the continuous scanning of the environment while the vehicle is in motion. These distortions arise from changes in the ego vehicle's position and orientation between consecutive frames, leading to discrepancies in the alignment of the point cloud data. To simulate these distortions, we implemented a segmented distortion method that differentiates the LiDAR scan area into N sectors based on their polar angles, where N=100 in our experiments.

Let the ego vehicle's position in the current time frame be \(\mathbf{P}_{t}\) and its position in the next time frame be \(\mathbf{P}_{t+1}\). The transformation matrix \(\mathbf{T}_{t}^{t+1}\) between these two frames is defined as:
\begin{equation}
    \label{Pt}
    \mathbf{P}_{t+1} = \mathbf{T}_{t}^{t+1} \cdot \mathbf{P}_{t}.
\end{equation}

The unit transformation matrix between each of the \(N\) sectors, \(\Delta \mathbf{T}_{u}\), is calculated by:
\begin{equation}
    \mathbf{T}_{t}^{t+1} = \sum_{n=0}^{N} \Delta \mathbf{T}_{u}, n \in N.
\end{equation}




Define the original undistorted points in sector $n$ as $\mathbf{p}_{n}'$, then, the transformation matrix for that sector can be computed as:
\begin{equation}
        \Delta \mathbf{T}_{n} = \sum_{n=0}^{n} \Delta \mathbf{T}_{u}.
\end{equation}

The distorted points $\mathbf{p}_{n}$ at sector $n$ is then computed as: 
\begin{equation}
    \label{p1}
    \mathbf{p}_{n} = \Delta \mathbf{T}_{n} \cdot \mathbf{p}_{n}'.
\end{equation}

By combining Equ. \ref{Pt} $\sim$ \ref{p1}, all the distorted LiDAR points $\mathbf{p}_{N}$ can be computed as the function of all the original undistorted LiDAR points $\mathbf{p}_{N}'$, ego vehicle position $\mathbf{P}_{t}$ at the current time frame and its position $\mathbf{P}_{t+1}$ at the next time frame shown in: 
\begin{equation}
    \mathbf{p}_{N} = f(\mathbf{p}_{N}', \mathbf{P}_{t}, \mathbf{P}_{t+1})
\end{equation}
This segmented approach ensures that distortions are applied locally, effectively mimicking the effects of vehicle dynamics on point cloud data. 
To illustrate the impact of these distortions, visualizations comparing the original undistorted LiDAR data with the distorted versions are shown in Fig.~\ref{fig:Motion_distortion}.

\begin{figure*}
\centering
\includegraphics[width=0.68\textwidth]{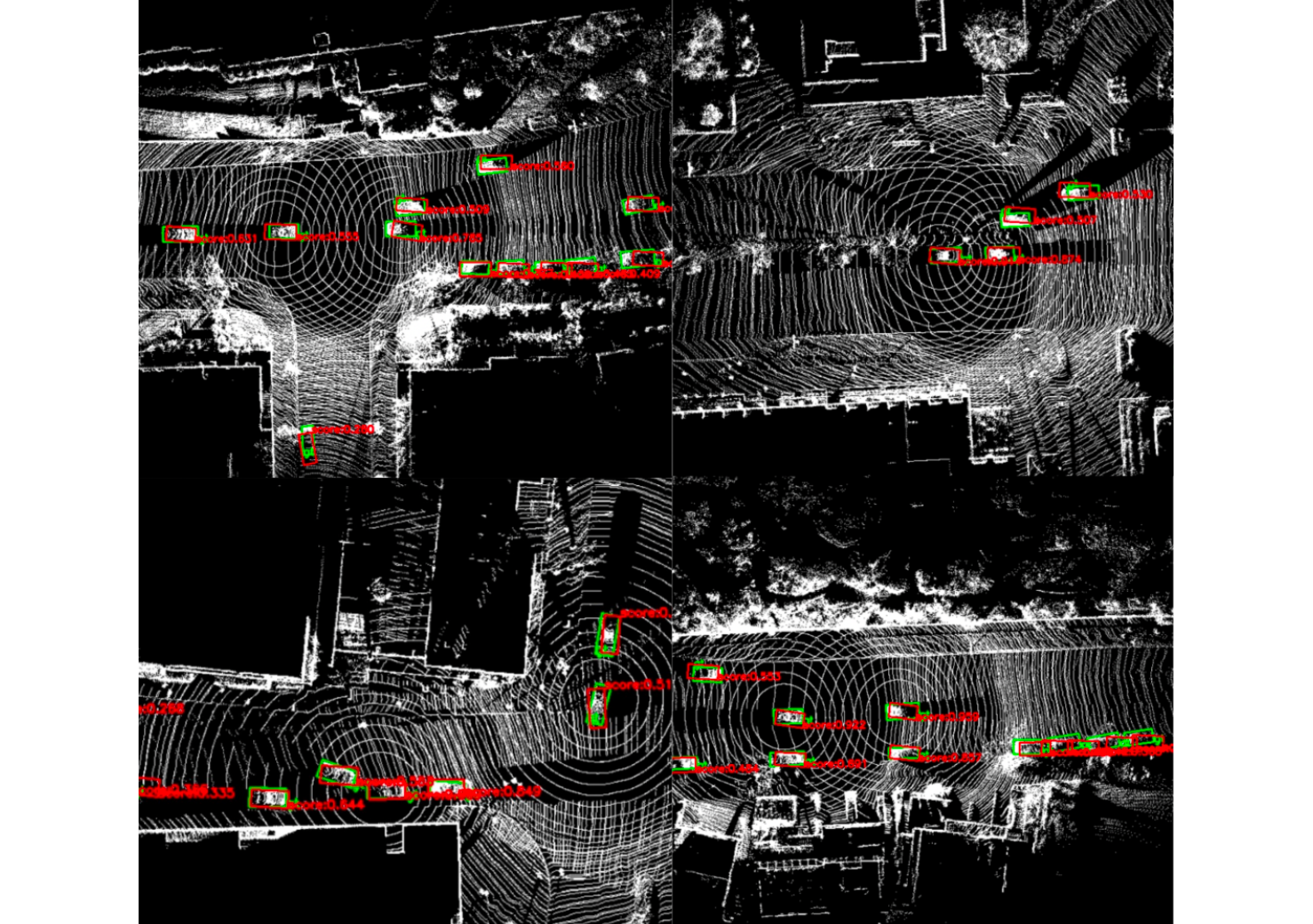}
\caption{
\textbf{Visual representation of detection results using our aligned camera features.} The figure illustrates the effectiveness of our proposed method in dense crowd multi-agent perception scenarios. 
}
\label{fig:camera_result}
\end{figure*}

\section{Time Synchronization}
\begin{figure*}
\centering
\includegraphics[width=0.98\textwidth]{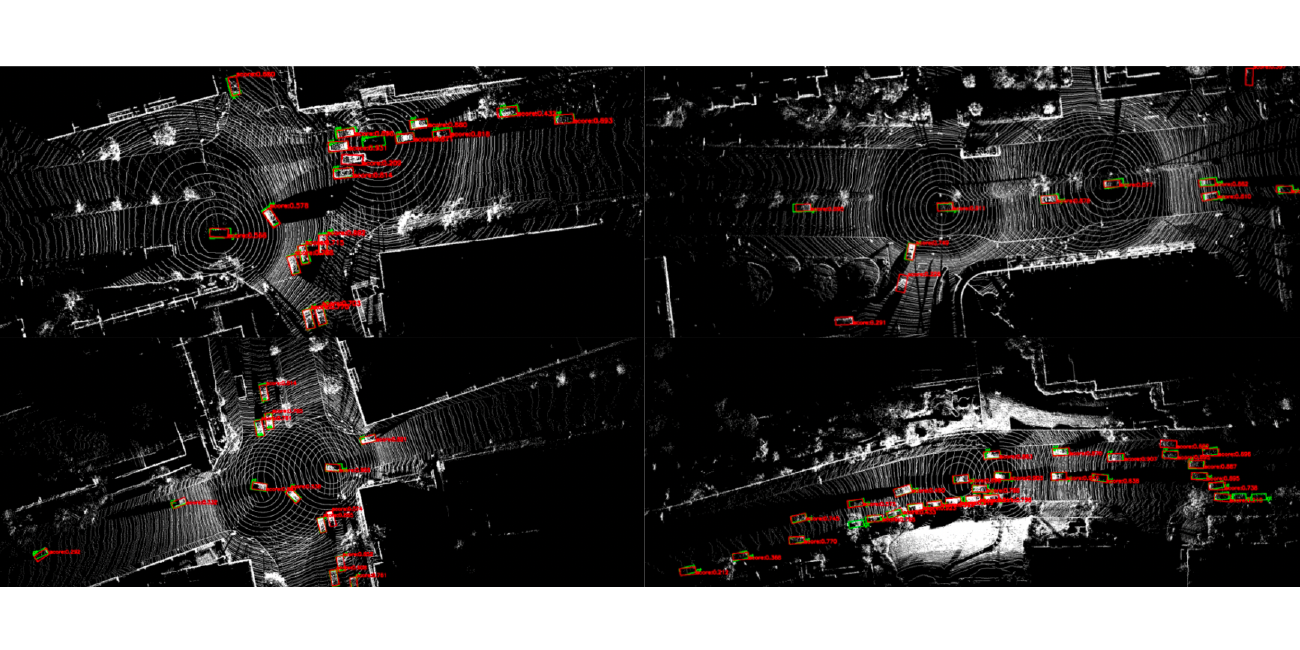}
\caption{
\textbf{Visual representation of detection results using both camera and LiDAR data.} The figure shows the detected objects where LiDAR data is used to complement camera-based processing for improved detection accuracy.
}
\label{fig:camera_lidar_result}
\end{figure*}

Time synchronization is a fundamental challenge in multi-sensor systems, particularly in V2X (vehicle-to-everything) scenarios, where data from heterogeneous sensors like LiDAR and cameras must be temporally aligned to ensure accurate perception. This synchronization is critical for fusing sensor measurements effectively, as even small temporal misalignments can lead to errors in object detection, tracking, and prediction. However, achieving perfect synchronization is often difficult due to several factors. First, hardware-level limitations, such as the inability of certain sensors to be hardware-triggered by a common clock (e.g., GPS signals), introduce delays. Second, software-based synchronization methods, while commonly used, rely on timestamp alignment and are susceptible to triggering errors due to network latency or processing delays. Finally, the inherent heterogeneity of sensors—each with unique data capture rates, triggering mechanisms, and latency—compounds these challenges, leading to asynchronous data collection. In multi-agent systems, this issue becomes even more pronounced as multiple vehicles and infrastructure units must synchronize their data across a shared communication framework.
To replicate these real-world challenges, we simulate asynchronous triggering between LiDAR and camera sensors in both infrastructure and vehicle sensor suites. In our simulation, we introduce a maximum temporal discrepancy of 0.1 seconds between the trigger times of the LiDAR and multiple camera sensors. This value reflects realistic temporal misalignments observed in V2X systems due to the aforementioned factors. By introducing this time synchronization noise, we aim to model the inherent complexities of multi-sensor and multi-agent systems, providing a testbed for evaluating the robustness of perception algorithms under realistic temporal discrepancies.

\section{Systematic Error}
\label{sec:systematic_error}

In real V2X systems, sensors may experience fixed misalignment errors due to external environmental interference~\cite{zhang2016systematic}. Factors such as temperature changes, moisture, and corrosion can cause sensors to shift to incorrect positions, resulting in persistent offsets that affect detection accuracy. To emulate these systematic errors, we apply fixed translation and rotation errors to the LiDAR point cloud data and camera images.

The fixed translation and rotation errors are applied by modifying the transformation matrix \( T_s^{'} \). Specifically,
 fixed rotation errors in yaw, pitch, and roll are uniformly sampled from \([-0.1^\circ, 0.1^\circ]\).
Fixed translation errors along the \( x, y, z \) axes (\( t_x, t_y, t_z \)) are uniformly sampled from \([-0.1\,\text{m}, 0.1\,\text{m}]\). The errors are added to reflect potential noise caused by long-term sensor corrosion, mechanical wear, or environmental factors such as temperature variations.

For camera images, fixed translation errors are simulated in the image plane. The errors along the horizontal (\( u \)) and vertical (\( v \)) axes (\( \Delta u, \Delta v \)) are uniformly sampled from a range equivalent to 1.5\% of the image resolution. 

The translation is applied using an affine transformation:
\begin{align}
    \mathbf{I}^{'}(x,y) = \mathbf{A}_s \cdot \mathbf{I}(x,y)
\end{align}
where:
\begin{align*}
\mathbf{A}_s = \begin{bmatrix}
1 & 0 & \Delta u \\
0 & 1 & \Delta v \\
0 & 0 & 1
\end{bmatrix}
\end{align*}
and $\mathbf{I}(x,y)$ and $\mathbf{I}^{'}(x,y)$ represents the image pixels before 
 and after transformation.

This systematic error simulation captures the realistic effects of fixed misalignment errors on both LiDAR and camera data.

\section{Qualitative Detection Results}
\label{sec:detection_result}

To demonstrate the effectiveness and resilience of our proposed network, we present qualitative results across various challenging detection scenarios in the V2X-Real dataset, as shown in Fig. \ref{fig:camera_result} and Fig. \ref{fig:camera_lidar_result}. In Fig. \ref{fig:camera_result}, the detection results of our aligned camera in densely crowded areas are highlighted, showcasing the network's ability to accurately detect occluded and distant objects. Furthermore, Fig. \ref{fig:camera_lidar_result} displays the fused camera and LiDAR detection outcomes in diverse scenarios, further emphasizing the robustness and accuracy of our approach.


\end{document}